\documentclass[preprint,12pt]{elsarticle}

\usepackage{amssymb}
\usepackage{amsmath}

\usepackage{graphicx}%
\usepackage{multirow}%
\usepackage{amsmath,amssymb,amsfonts}%
\usepackage{amsthm}%
\usepackage{mathrsfs}%
\usepackage[title]{appendix}%
\usepackage{xcolor}%
\usepackage{textcomp}%
\usepackage{manyfoot}%
\usepackage{booktabs}%
\usepackage{algorithm}%
\usepackage{algorithmicx}%
\usepackage{algpseudocode}%
\usepackage{listings}%

\usepackage{hyperref}
\usepackage{url}
\usepackage{svg}
\usepackage{algorithm}
\usepackage{algpseudocode}
\usepackage{amsmath}
\usepackage{amsthm}
\usepackage{makecell}
\usepackage{multirow} 
\usepackage{booktabs}
\usepackage{amsthm}
\usepackage{amssymb}
\usepackage{float}
\usepackage{footnote}

\newtheorem{definition}{Definition}
\newtheorem{proposition}{Proposition}

\journal{Nuclear Physics B}

\begin{document}

\begin{frontmatter}



\title{A Constraint-Preserving Neural Network Approach for Mean-Field Games Equilibria} 


\author[1,3]{Jinwei Liu}\ead{liujinwei@buaa.edu.cn} 

\author[1,3]{Lu Ren}\ead{urrenlu@buaa.edu.cn}

\author[2,3,4,5]{Wang Yao\corref{cor1}}\ead{yaowang@buaa.edu.cn}

\author[1,3,4,5]{Xiao Zhang\corref{cor1}}\ead{Xiao.zh@buaa.edu.cn}

\affiliation[1]{organization={School of Mathematical Sciences},
            addressline={Beihang University}, 
            city={Beijing 100191},
            country={China}}
\affiliation[2]{organization={School of Artificial Intelligence},
            addressline={Beihang University}, 
            city={Beijing 100191},
            country={China}}
\affiliation[3]{organization={Key Laboratory of Mathematics, Information and Behavioral Semantics (LMIB)},
            addressline={Beihang University}, 
            city={Beijing 100191},
            country={China}}
\affiliation[4]{organization={Hangzhou International Innovation Institute of Beihang University},
            city={Hangzhou 311115},
            country={China}}
\affiliation[5]{organization={Zhongguancun Laboratory},
            city={Beijing 100094},
            country={China}}
\cortext[cor1]{These authors contributed equally to this work and should be considered co-corresponding authors.}

\begin{abstract}

Neural network-based methods have demonstrated effectiveness in solving high-dimensional Mean-Field Games (MFG) equilibria, yet ensuring mathematically consistent density-coupled evolution remains a major challenge. This paper proposes the NF-MKV Net, a neural network approach that integrates process-regularized normalizing flow (NF) with state-policy-connected time-series neural networks to solve MKV FBSDEs and their associated fixed-point formulations of MFG equilibria. The method first reformulates MFG equilibria as MKV FBSDEs, embedding density evolution into equation coefficients within a probabilistic framework. Neural networks are then employed to approximate value functions and their gradients. To enforce volumetric invariance and temporal continuity, NF architectures impose loss constraints on each density transfer function. Theoretical analysis establishes the algorithm’s validity, while numerical experiments across various scenarios including traffic flow, crowd motion, and obstacle avoidance demonstrate its capability in maintaining density consistency and temporal smoothness.
\end{abstract}


\begin{highlights}
\item NF-MKV Net integrates NF and NNs for high-dimensional MFG solutions.
\item Process-regularized NF ensures density transformation consistency across time steps.
\item State-policy-connected time series neural networks approximates time-consistent solutions.
\item Validated on Traffic Flow, Crowd Motion, and Obstacle Tasks, distinguishing OT.
\item Outperforms existing methods with Volumetric and Time-Continuity constraints.
\end{highlights}

\begin{keyword}
Mean-Field Games\sep normalizing flow\sep McKean-Vlasov forward-backward stochastic differential equations\sep mathematical constraints neural network



\end{keyword}
\end{frontmatter}



\section{Introduction}
Mean-Field Games (MFG), introduced independently by Lasry-Lions \cite{LASRY2006619, LASRY2006620, lasry2007mean} and Huang-Malham{\'e}-Caines \cite{cis/1183728987, 4303232}, offer a robust framework for solving large-scale multi-agent problems. Due to its wide application in fields such as autonomous driving \cite{7364206, 9061051}, social networks \cite{WOS:000352223501068}, crowd management \cite{LACHAPELLE20111572}, and power systems \cite{7931716}, how to solve for the equilibria of MFG has become one of the hot topics in current research.

Most algorithms are to seek the equilibria of MFG by solving the solutions to the Hamilton-Jacobi-Bellman (HJB) and Fokker-Planck-Kolmogorov (FPK) equations. Although solution algorithms based on traditional numerical methods, such as the finite element and finite difference \cite{doi:10.1142/S0218202512500224,Achdou2013aaa,doi:10.1137/120882421,osborne2025finite}, have also been studied, solution algorithms based on neural networks are attracting more and more attention, due to their ability to effectively overcome the curse of dimensionality when dealing with high-dimensional problems. For instance, HJB-FPK equations can be reformulated as a Generative Adversarial Network (GAN) \cite{lin2021alternating} training problem. A Lagrangian-based approach \cite{ruthotto2020machine} has been proposed to approximate agent states through sampling and formulate neural networks to solve MFG trace problems. The self-attention mechanism has been employed to enhance the scalability of the solution \cite{YIN2024106002} for generalized initial and terminal value function. Additionally, reinforcement learning and neural networks \cite{chen2023hybrid} are used to model distributions and address value functions. Existing neural network-based methods have demonstrated good performance in solving problems and have already been applied in specific scenarios. However, these algorithms typically replace the continuous density in MFG equilibria seeking by sampling N agents’ states as $N$-dimensional vectors (with $N$ large enough), which will naturally lose accuracy and cannot guarantee the maintenance of normative constraints during the density evolution process.

By leveraging MKV FBSDEs to address MFG equilibria \cite{carmona2018probabilistic, 10.1214/14-AAP1020, doi:10.1137/120883499}, stochastic process perspective is introduced to explore numerical methods for solving MFG. A simplified MFG model \cite{achdou2015system} is proposed for pedestrian dynamics and demonstrated through numerical simulations. By solving asymmetric Riccati differential equations and establishing sufficient conditions for the existence and uniqueness of optimal solutions, multi-group MFG can also be addressed from a stochastic process perspective \cite{ren2024hierarchical}. MKV FBSDEs are based on stochastic dynamics modeling, ensuring the continuity of agents transitions. These methods have been the subject of in-depth theoretical research and enjoy widespread application in the context of linear-quadratic MFG such as opinion dynamics and financial analysis. However, to our knowledge, except for an intelligent method that transforms mean field problems in specific cases into optimal transport problems for solution, which will be discussed in detail in Section 2, methods based on solving the MKV FBSDEs have not yet been used to solve the equilibria of MFG other than linear quadratic ones.

To address the challenges in solving high-dimensional nonlinear MFG with density-distribution coupling, we proposes NF-MKV Net. Solving MFG equilibria can be translated into addressing equivalent stochastic fixed-point problems that incorporate density flows. NF-MKV Net can solve the MKV FBSDEs problem by coupling the process-regularized NF with state-policy-connected time-series neural networks, which ensures maintaining density consistency and temporal smoothness.

The new features of this paper can be summarized as follows:
\begin{itemize}
    \item \textbf{NF-MKV Net} is proposed to solve MFG from a stochastic process perspective. By integrating process-regularized NF and neural network into a coupled approach with alternating training, the approach can overcome the challenges of high-dimensional, density-coupling and time-continuous.

    \item \textbf{Process-regularized NF frameworks} are designed to model probability measure flows by enforcing loss constraints on each density transfer function, thereby maintaining volumetric invariance in density transformations at each time step.
    
    \item \textbf{State-policy-connected time-series Neural Networks}, grounded in MKV FBSDEs, model the relationships between time-step value functions and approximate their gradients, enabling solutions that are consistent over time.

    \item The proposed approach is demonstrated to be applicable in diverse scenarios, including density-coupling, low- and high-dimensional, obstacle avoidance and terminal value function constraints problems, maintaining density consistency and temporal smoothness.
    
\end{itemize}

The remainder of this paper is organized as follows: Section \ref{S2} provides the preliminaries of the problem settings. Section \ref{S3} proposes the NF-MKV Net approach and gives the framework. Section \ref{S4} provides the theory analysis of the approach with computational error and complexity. Section \ref{S5} shows the numerical results and the conclusion is given in Section \ref{S6}.

\section{Preliminaries}
\label{S2}
\subsection{Formulations of MFG: HJB-FPK Equations and MKV FBSDEs.}
Without lose of the generality, we present the mathematical formulations of MFG defined by HJB-FPK equations and by MKV FBSDEs in the $d-$dimensional $\mathbb{R}^d$ space, and demonstrate their equivalence.

\begin{definition}[MFG defined by HJB-FPK \cite{lasry2007mean}]

\begin{equation}\begin{cases}
\textnormal{(HJB)} \; -\partial_t u(t,x) - \frac{1}{2}\text{Tr}\left(\sigma(x)\sigma(x)^\top D^2u(t,x)\right) + H(x,Du(t,x),\mu_t)=0 ,\\\textnormal{(FPK)}\;  
\partial_t\mu_t(x) - \frac{1}{2}\text{Tr}\left(\sigma(x)\sigma(x)^{\top}D^2\mu_t(x)\right) - \nabla\cdot(\mu_t(x)b(x,\alpha_t^*,\mu_t)) = 0 ,\\
\mu_{t}|_{t=0}(x)=\mu_0 (x),\quad u(T,x)=g(x),\quad \int_\Omega \mu_t (x) \mathrm{d}x = 1\quad  t\in [0,T],
\end{cases}\label{mfg}\end{equation}
where $u:\mathbb{R}^d\times [0,T]\rightarrow \mathbb{R}$  is the value function to guide the agents to make decisions;  $H(x,Du(t,x),\mu_t)= \min _{\alpha_t \in \mathbb{A}} [{f(x,\mu_t)+b\cdot Du(t,x)}]$ is the Hamiltonian, which describes the physics energy of the system; $D$ is a differential operator; $\mu_t(x)\in \mathcal{L}(\mathbb{R}^d)$ is the distribution of agents at time $t$, $f:\mathbb{R}^d\times \mathcal{L}(\mathbb{R}^d)\rightarrow \mathbb{R}$ denotes the loss during process; $\mu_0$ is the initial distribution and $g:\mathbb{R}^d\rightarrow \mathbb{R}$ is the terminal value function, guiding the agents to the terminal condition.
\end{definition}

From the perspective of stochastic processes purpose, we define a complete filtered probability space $(\Omega,\mathcal{F},\mathbb{F}=\left(\mathcal{F}_t)_{0\leq t\leq T},\mathbb{P}\right)$, where the filtration $\mathbb{F}$ is adapted to a $d-$dimensional Wiener process $\mathbf{W}=(W_t)_{0\leq t\leq T}$ with respect to an initial condition $\xi\in L^2(\Omega,\mathcal{F}_0,\mathbb{P};\mathbb{R}^d)$. This MFG problem can be described as:
\begin{definition}[MFG defined by Stochastic Process \cite{carmona2018probabilistic}]
~

(i) For each fixed deterministic flow $\boldsymbol{\mu} = (\mu_t)_{0\leq t\leq T}$ of probability measures on $\mathbb{R}^d$, we solve the standard stochastic control problem

\begin{equation}
\inf_{\alpha\in \mathbb{A}}J^{\boldsymbol{\mu}} (\alpha) \quad \text{with}\quad J^{\boldsymbol{\mu}} (\alpha)=\mathbb{E}\left[\int_0^T f(t,X^\alpha _t,\mu_t,\alpha_t)\mathrm{d}t+g(X_T^\alpha,\mu_T)\right],
\label{con}
\end{equation}
subject to
\begin{equation}
    \left\{
    \begin{array}{l}
         dX_t^\alpha =b(t,X_T^\alpha,\mu_t,\alpha_t)\mathrm{d}t+\sigma(t,X_T^\alpha,\mu_t,\alpha_t)\mathrm{d}W_t\quad t\in[0,T],  \\
         X_0^\alpha =\xi,
    \end{array}
    \right.\nonumber
\end{equation}
where $X_t^\alpha$ is the statement of agents at time $t$ with control $\alpha$; $\alpha$ is the control of representative agent; $\mathbb{A}$ is the set of control and $J$ is the loss of MFG system, including the process loss $f$ and terminal value function $g$.

(ii) Find a flow $\boldsymbol{\mu}=(\mu_t)_{0\leq t\leq T}$ such that $\mathcal{L}(\hat{X}_t^{\boldsymbol{\mu}})=\mu_t$ for all $t\in[0,T]$, if $(\hat{X}^{\boldsymbol{\mu}}_t)_{0\leq t \leq T}$ is a solution of the above optimal control problem.
\end{definition}

Under Assumptions \cite{carmona2018probabilistic}, for any initial condition $\xi\in L^2(\Omega,\mathcal{F}_0,\mathbb{P};\mathbb{R}^d)$, the MKV FBSDEs

\begin{equation}
    \begin{cases}
    \begin{aligned}
    dX_t=&b(t, X_t, \mathcal{L}(X_t), \hat{\alpha}(t, X_t, \mathcal{L}(X_t), \sigma(t, X_t, \mathcal{L}(X_t))^{-1 \dagger} Z_t)) \mathrm{d}t \\ &+\sigma(t, X_t, \mathcal{L}(X_t)) d W_t, \\d Y_t=&-f(t, X_t, \mathcal{L}(X_t), \hat{\alpha}(t, X_t, \mathcal{L}(X_t), \sigma(t, X_t, \mathcal{L}(X_t))^{-1 \dagger} Z_t)) d t\\ &+Z_t \cdot d W_t,
    \end{aligned}
\end{cases}
\label{fbsde}
\end{equation}
for $t\in [0,T]$, with $Y_T=g(X_T,\mathcal{L}(X_T))$ as terminal value function, is \textbf{solvable}, where $Y_t$ is the value function of representative agent, $Z_t$ is adjoint process of $Y_t$, $\mathcal{L}(X_t)$ is the marginal distribution of $X_t$ and $(-1\dagger)$ denotes the adjoint operator corresponding to the inverse of the differential operator.

\begin{proposition} [Equivalent \cite{carmona2018probabilistic}]
The flow $(\mathcal{L}(X_t))_{0\leq t\leq T}$ given by the marginal distributions of the forward component of any solution is an equilibria of the MFG problem \ref{mfg} associated with the stochastic control problem \ref{con}.
\end{proposition}

\subsection{Normalizing Flow}
NF enables exact computation of data likelihood through a series of invertible mappings \cite{tabak2010density,rezende2015variational} and are widely used in stochastic system modeling \cite{9089305,JMLR19-1028,21M1450604}. A key feature of NF is its use of arbitrary bijective functions, implemented through stacked reversible transformations. The flow model $R(x)$ consists of a series of reversible flows, expressed as $R(x) = r_1 \circ r_2 \circ \cdots \circ r_L(x),\; L\in N^*$, where each $r_i$ has a tractable inverse and Jacobian determinant.

In an NF model, if each $r_i$ for $i=1,2,\cdots,N$ is differentiable and reversible functions, they are typically expressed as
\begin{equation}
\begin{array}{ll}
    \text{(Normalizing)}\quad \mathbf{r}=r_1\circ r_2\circ\cdots \circ r_N , \\ \mathbf{p}_{\mu_0}(X)=\quad \mathbf{p}_{\mu_T}(\mathbf{r}(X)) | \det D\mathbf{r}(X)|,\\
    \text{(Construct)}\quad \mathbf{s}=s_N\circ s_{N-1}\circ\cdots \circ s_1 ,\\ \mathbf{p}_{\mu_T}(X)=\quad \mathbf{p}_{\mu_0}(\mathbf{s}(X)) | \det D\mathbf{r}(X)|^{-1}.
\end{array}\nonumber
\end{equation}
During training, each $r_i(x)$ is represented as a neural network $r_i(x; \phi)$. Multiple $r_i$ functions are combined to obtain the desired function $f$ and minimize the negative log-likelihood loss between the final estimated density and the dataset, which is expressed as
\begin{equation}
    L(x)=-\log \mathbf{p}_{\mu_T}(x)=-\log \mathbf{p}_{\mu_0}(\mathbf{r}^{-1}(x))-\log |\det D\mathbf{r}^{-1}(x)|.\nonumber
\end{equation}

The connection between MFG and NF offers inherent advantages. In MFG, the initial distribution is often represented in a simple analytical form, similar to NF, where an initial simple distribution transforms into a more complex one for density estimation. Additionally, the volume-preserving property of NF aligns with the MFG requirement $\int_\Omega \mu(x,t) \mathrm{d}x = 1$ for $t\in [0,T]$, as proposed by Lasry-Lions \cite{lasry2007mean} in equations \ref{mfg}.

\subsection{Connections and Differences between MFG and OT}
The OT problem can be defined as follows:
\begin{definition}[Optimal Transport \cite{santambrogio2015optimal}]
    Given probability measures $\mu \in P(\mathcal{X})$ and $\nu \in P(\mathcal{Y})$, consider a mapping $T : \mathcal{X} \rightarrow \mathcal{Y}$ such that $T_{\#}\mu = \nu$ (i.e., $T$ pushes $\mu$ forward to $\nu$). The OT problem seeks to minimize the total transport cost
    \begin{equation*}\inf_{T_\#\mu=\nu}\int_{\mathcal{X}} c(x,T(x))d\mu(x),\end{equation*}
where $c(x,y)$ is the cost function for transporting $x$ to $y$.

In the Kantorovich formulation, by considering couplings $\gamma \in \Pi(\mu,\nu)$ (i.e., joint distributions with marginals $\mu$ and $\nu$), the OT problem is given by
\begin{equation*}
\inf_{\gamma \in \Pi(\mu,\nu)} \int_{X \times Y} c(x,y) d\gamma(x,y).
\end{equation*}
\end{definition}
MFG and OT are similar, but they differ significantly. Both MFG and OT involve the evolution of densities. In MFG, the evolution of the population distribution described by the FPK equation can be regarded as the push-forward of probability measures in the OT process. OT incorporating process-based losses has been employed to solve MFG equilibria \cite{huang2023bridging}. However, OT demands a clear initial specification of the terminal density distribution, whereas MFG guides the evolution of the density through terminal value function without knowing the terminal distribution in advance.

\section{Methodology: NF-MKV Net}
\label{S3}
We propose NF-MKV Net, an approach trained alternately with NF and MKV FBSDEs, for solving MFG equilibria, inspired by the Deep BSDE approach \cite{pnas1718942115} for solving the single-agent HJB equations.

\begin{figure}[ht]
\begin{center}
\includegraphics[width=\linewidth]{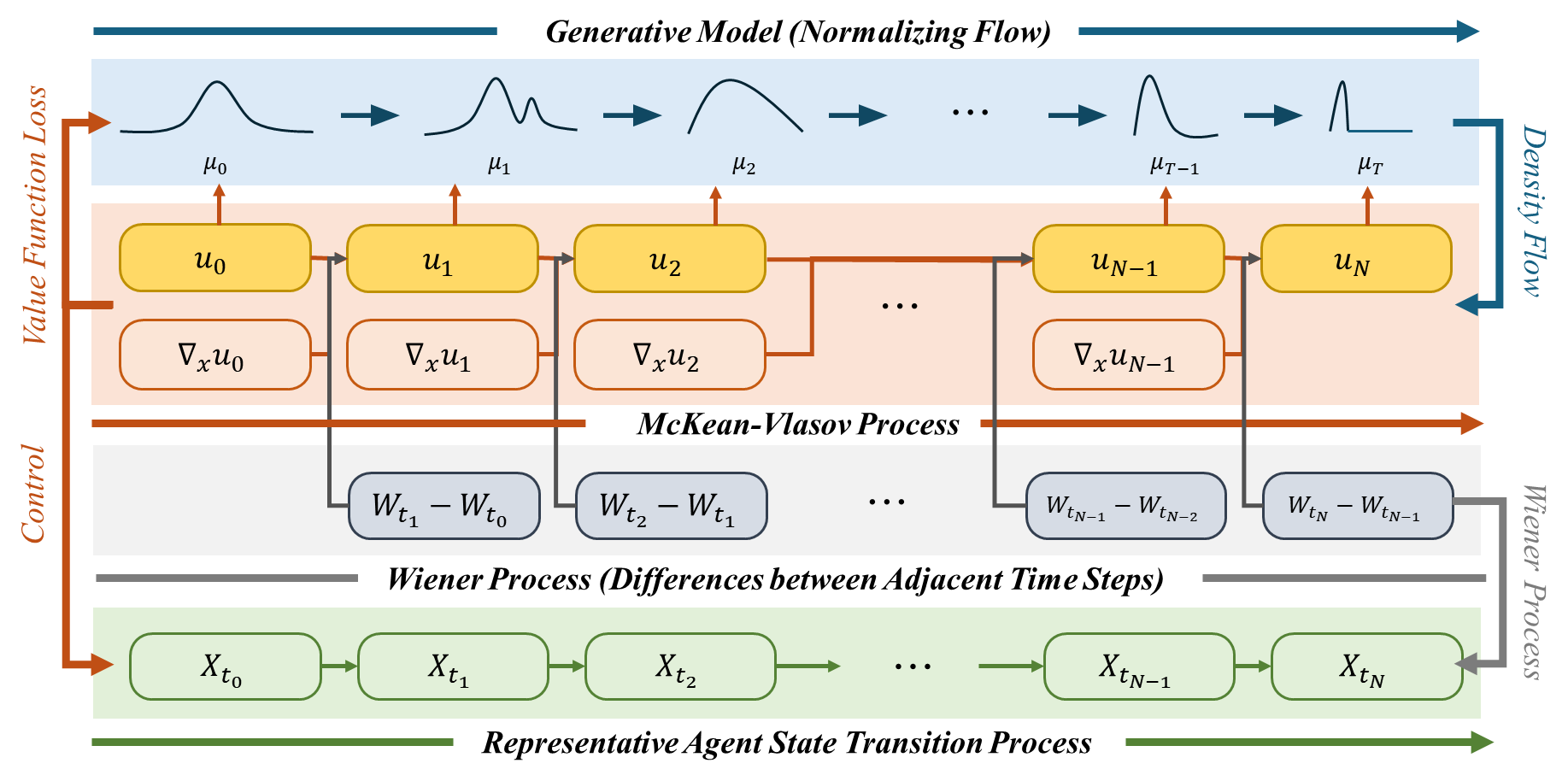}
\end{center}
\caption{Approach Diagram of the NF-MKV Net}
\label{fig-1}
\end{figure}

Figure \ref{fig-1} illustrates the NF-MKV Net approach. The NF-MKV Net consists of two components: a value function gradient neural network (red part) that characterizes the evolution of the representative agent's value function, and a NF structure (blue part) designed to approximate the marginal distribution of the representative agent at each time. NF-MKV Net can capture both optimization and interaction components within a single coupled FBSDE, eliminating the need for separate references to the HJB and FPK equations. 

The value function constrains the neural network generating the flow via the HJB equation and terminal value function, while its gradient update depends on the current marginal density flow (\ref{31}). NF generates flows through Neural Networks, constraining each density transfer function to define distributions at specific times. The density flow from NF is coupled with the MFG value function (\ref{s32}). By integrating process-regularized NF and neural network into a coupled approach with alternating training, the approach can overcome the challenges of high-dimensional, density-coupling and time continuity (\ref{33}).


\subsection{Modeling Value Function with MKV FBSDEs}
\label{31}
In modeling the value function network, our approach applies discretization to the value function equation corresponding to the MKV FBSDEs in \ref{311}. Subsequently, we leverage the relationship between the value function at each time steps in the discretized equation, combining it with a neural network to model the gradient of the value function in \ref{322}.

\subsubsection{Transformation of the value function equation}
\label{311}
We consider a general class of MFG problem associated with the stochastic control problem \ref{con}, the relative FBSDEs can be represented as in equations \ref{fbsde}, with initial condition $\xi\in L^2(\Omega,\mathcal{F}_0,\mathbb{P};\mathbb{R}^d)$ and terminal value function $Y_T=g(X_T,\mathcal{L}(X_T))$.

Then the solution of equation \ref{mfg} satisfies the following FBSDE
\begin{equation}
\begin{aligned}
    &u(x,t)-u(x,0)= \\ &-\int_0^t f(s, X_s, \mathcal{L}(X_s), \hat{\alpha}(s, X_s, \mathcal{L}(X_s), \sigma(t, X_s, \mathcal{L}(X_s))^{-1 \dagger} Z_s)) \mathrm{d}s  + \int_0^t Z_s \mathrm{d}W_s.
\end{aligned}
    \label{ude}
\end{equation}

We apply a temporal discretization to equation \ref{fbsde}. Given a partition of the time interval $[0,T]:0=t_0<t_1 <\cdots<t_N=T$, we consider the Euler-Maruyama method for $n=1, \cdots , N-1$,
\begin{equation}
    X_{t_{n+1}} - X_{t_{n}} \approx b(t,X_T^\alpha,\mu_t,\alpha_t)\Delta t_n +\sigma(t,X_T^\alpha,\mu_t,\alpha_t)\Delta W_n,\label{eux}\nonumber
\end{equation}
and
\begin{equation}
    u(x,t_{n+1})-u(x,t_n) \approx -f(s, X_s, \mathcal{L}(X_s), \hat{\alpha}_t(\cdot )^{-1 \dagger} Z_s)) \Delta t_n+[\partial_x u(x,t)]^T\sigma \Delta W_n \label{euler},
\end{equation}
where
\begin{equation}
    \Delta t_n=t_{n+1}-t_n,\quad \Delta W_n=W_{t_{n+1}}-W_{t_n}.\nonumber
\end{equation}

\subsubsection{Construction of the value function neural network}
\label{322}
The key to modeling the above FBSDEs is approximating the value function $x \mapsto u(0,x)$ at $t = t_0$, such that $u(0,x) \approx u(0,x | \theta_0)$, and approximating the function $x \mapsto [\partial_x u(x,t)]^T\sigma$ at each time step $t = t_n$ using a multi-layer feedforward neural network
\begin{equation}
x \mapsto [\partial_x u(x,t)]^T\sigma \approx [\partial_x u(x,t) | \theta_n]^T\sigma. \nonumber
\end{equation}
This represents the adjoint variable $Z_t$ of the value function of the representative agent, expressed as the product of the gradient and the random variable in the MFG optimization problem.

All value functions are connected by summing equation \ref{euler} over $t$. The network takes the generated density flows $\boldsymbol{\mu}$ and $W_{t_n}$ as inputs and produces the final output $\hat{u}$, approximating $u(x, T) = g(x, \mu(\cdot, T))$. This approximation defines the expected loss function by comparing the difference in terminal value function at sampled points under the current terminal distribution $\mu_T$. For $\{x_i\}_{i=1}^N = z \sim \mu_T$,
\begin{equation}
l_{\text{MKV}}=|\hat{u}(\theta,z))-g(z,\mu_T)|^2 = \frac{1}{N} \sum\nolimits_{i=1}^N |\hat{u}(\theta,x_i))-g(x_i,\mu_T)|^2.\label{nll1}
\end{equation}

We reformulate MFG as MKV FBSDEs in equation \ref{ude} and discretize time to establish the relationship between value functions at each time step $t$ in equation \ref{euler}, connecting them via the adjoint variable $Z_t$. Next, we parameterize $Z_t$ and $u_0$, using these relationships to link $u_T$ with the terminal value function $g(x, \mu(\cdot, T))$ in equation \ref{nll1} to minimize the loss of terminal value function.

\subsection{Modeling density distribution with NF}
\label{s32}
In modeling the flow of marginal distributions by NF, NF-MKV Net aligns the transition functions in the NF with discrete time steps. Additionally, we modify the terminal matching term to correspond to the terminal loss of value function in the MFG formulation in \ref{321}. Subsequently, we utilize the state variables of the representative agent at each time step, along with the terminal matching condition, to model the marginal distribution of the representative agent using NF in \ref{s322}.
\subsubsection{Refinement of the NF architecture}
\label{321}
\textbf{a. Align with the MFG framework.} Direct application of the NF method to estimate the marginal distribution in MFG problems presents certain challenges. NF methods typically focus on density estimation results. In contrast, our approach emphasizes the evolution of density flows in NF, constraining each layer to align with the density evolution in MFG. Existing NF-based approaches typically rely on the known density distribution of terminal states, using metrics like KL divergence or MMD as the loss function.

To address this issue, we propose enhancing the network architecture of the NF model. After sampling the density distribution of terminal states, we substitute the form of the terminal value function, compute the loss
\begin{equation*}
    l=\|g(z,\mu_T)\|^2\quad \text{for} \quad z \sim \mu_T,
\end{equation*}
and use it as the optimization objective to enhance the NF network architecture, aligning with the loss function optimization objectives in MFG.

\textbf{b. Correspond to discrete time steps. }Discretizing the NF construction process shows that each function corresponds to an Euler time discretization. Therefore, each sub-function $r_i$ transforms the group density $\mu_i$ in MFG into $\mu_{i+1}$. This series of reversible flows represents the time evolution process, leading to
\begin{equation}
    \mu_0\xrightarrow{\mathbf{r}(x)} \mu_T \quad \Leftrightarrow \quad \mu_0\xrightarrow{r_{t_1}(x)} \mu_{t_1}\xrightarrow{r_{t_1}(x)} \mu_{t_2}\xrightarrow{r_{t_2}(x)}\cdots \xrightarrow{r_{t_{N-1}}(x)} \mu_{t_N}.
    \label{mumu}\nonumber
\end{equation}

Additionally, since each sub-function is implemented as a neural network, the loss at each time step can be used to constrain and optimize the sub-functions. The density $\mu_t$ can be expressed as $\mu_{t_n} = \mathbf{r}_{1,2,\cdots,n} \circ \mu_0$, where $\mathbf{r}_{1,2,\cdots,n} = r_{t_n} \circ r_{t_{n-1}} \circ \cdots \circ r_{t_1}$.
\subsubsection{Construction of the marginal distribution NF}
\label{s322}
We model the system by approximating the density distribution at each layer
 \begin{equation} \mu_0 \mapsto \mathbf{r}_{1,2,\cdots ,n}(x) \circ \mu_0(x) \approx \mathbf{r}_{1,2,\cdots ,n}(x; \boldsymbol{\Phi}) \circ \mu_0 \end{equation} at each time step $t = t_n$. This is achieved through an NF model composed of multiple layers of Masked Autoregressive Flow (MAF) and Permute, parameterized by $\phi$.

\textbf{a. Computing process loss.} The first step in training the NF is computing the process loss $l_{\text{dis}}$. At each time step $t = t_n$, the state of a representative element, evolved according to the MKV FBSDEs, can be written as the marginal distribution $Law(X_{t_n})$, which is fitted using one of the transfer states of the regularized flow. In practice, a density function is fitted to the state quantities of all individuals representing the element, using the Negative Log-Likelihood (NLL) as the loss function 
\begin{equation}
\begin{aligned}
    l_{\text{dis}}=-\mathbb{E}_{X_{t_n}\sim\mathrm{Law}(X_{t_n})}\left[\log \mathbf{r}_{1,2,\cdots ,n}(x)\circ\mu_0(X_{t_n};\boldsymbol{\Phi})\right].
\end{aligned}
\label{ldis}
\end{equation}

\textbf{b. Satisfying terminal value function.} Additionally, the NF method must satisfy the terminal value function, so the loss of terminal value function $l_\text{T}$ is included in the loss calculation. The terminal state of the NF can be constrained by calculating the value of the terminal value function $g$ at the sampled point, under the terminal state $\mu_T(\Phi)$ of the density flow generated by the NF. The absolute value of the terminal value function, calculated with sampled points in $\mu_T(\Phi)$ generated by NF, is used to compute the loss of terminal value function $l_\text{T}$. For $\{x_i\}_{i=1}^N = z \sim \mu_T(\Phi)$
\begin{equation}
    l_{\textit{T}}=\|g(z,\mu_T(\Phi))\|^2=\frac{1}{N}\sum\nolimits _{i=1}^N \|g(x_i,\mu_T(\Phi))\|^2.\label{nll2}
\end{equation}

NF relies solely on the initial distribution $\mu_0$ and the loss of terminal value function $g$, while preserving density consistency. The losses $l_{\text{dis}}$ (equation \ref{ldis}) and $l_{\textit{T}}$ (equation \ref{nll2}) constrain the evolution of NF, ensuring that the flow density aligns with the control objectives.

\subsection{Coupling two processes}
\label{33}

In the NF-MKV Net approach, we separately model the state evolution of the representative agent and the marginal distribution. On the one hand, the state evolution of all representative agents leads to different marginal distributions, which in turn causes changes in the density distribution flow. On the other hand, after obtaining the optimal flow from the NF-MKV Net, the representative agent can directly substitute the optimal marginal distribution flow into the MKV FBSDEs to compute the optimal value function and control, thus solving the evolution process. 

Two processes can be coupled and trained alternately. As NF is a generative model, it can first generate a set of flow density evolution functions along with the corresponding density distributions at each time step. This generated set of density distributions is fixed as the marginal distribution to optimize the value function and gradient under the MKV FBSDEs framework. Once the optimal value function for this marginal density flow is obtained, it is fixed to update each $r_n$ and its corresponding $\mu_{t_n}$ in the NF evolution process. This continues until the optimal density flow under the current value function is achieved. This iterative coupled training continues until convergence. Algorithm \ref{alg1} presents the pseudo-code of the model.

\begin{algorithm}[H]
    \caption{NF-MKV Net}
    \label{alg1}
    \begin{algorithmic}
        \Require{$\sigma$ diffusion parameter, $g$ terminal value function, $H$ Hamiltonian, $f$ process loss, $\mu_0$ initial density}
        \Ensure{$\boldsymbol{\mu}=(\mu_t)_{0\leq t\leq T}$ density flow, $\mathbf{u}=(u_t)_{0\leq t\leq T}$, value function}
        \State $\mu_T$$\gets $$\arg \max_\mu g(x, \mu(x, T))$
        \State Generating NF $\{\mu_{t_n}(\phi_n)\}_{n=1}^N$ from $\mu_0$ to $\mu_T$
        \While{not converged}
        \State \textbf{Train} $u(0,x|\theta_0)$ and $[\partial_x u(x,t) | \theta_n]^T\sigma$ for $n=1,2,\cdots ,N$:
        \State Sample batch $\left(\{x_i\}_{i=1}^M,t_n\right)\sim \mu_{t_n}$ for $n=1,2,\cdots ,N$
        \State Sample Winner Process $\{W_{t_n}\}_{n=1}^N\sim \mathcal{N}(0,\sigma^2)$
        \State $l_{\textit{MKV}} \gets -\frac{1}{N}\sum\nolimits_{i=1}^N|g((x_i, T),\mu_T)-\hat{u}(\{(x_{i,n},t_n)\},\{W_{t_n}\}_{n=1}^N)|^2$
        \State Back-propagate the loss $l_{\textit{MKV}}$ to $\theta$ weights.
        \State \textbf{Train} $r_n(\phi_n)$ for $n=1,2,\cdots ,N$:
        \State Get trace from MKV $\left(\{x_i\}_{i=1}^M,t_n\right)$ for $n=1,2,\cdots ,N$\;
        \State $l_{\text{dis}}=-\mathbb{E}_{x_{t_n}\sim\mathrm{Law}(X_{t_n})}\left[\log \mathbf{r}_{1,2,\cdots ,n}(x)\circ\mu_0(X_{t_n};\boldsymbol{\Phi})\right]$
        \State Sample batch $ \{x_i\}_{i=1}^N \sim \mu_T(\phi_T)$
        \State $l_{\text{T}}\gets \frac{1}{N}\sum\nolimits _{i=1}^N \|g(x_i,\mu_T(\phi_T))\|^2$
        \State Back-propagate the loss $l_\text{NF}=l_{\text {dis}}+l_{\textit{T}}$ to $\phi$ weights.
        \EndWhile
    \end{algorithmic}
\end{algorithm}

\section{Analysis of Computational Error and Complexity}
\label{S4}
To better apply the algorithm, this paper provides a theoretical analysis of the NF-MKV Net approach from two aspects: computational error and computational complexity. Theoretical results are presented, and a comparison of computational complexity is made with existing methods for solving MFG equilibria.
\subsection{Analysis of Error Estimate}
The computational error of NF-MKV Net primarily arises from two sources: the error induced by time discretization (\ref{411}) and the density estimation error resulting from sampling the representative agent during the process (\ref{412}).
\subsubsection{Approximation Error of Time Discretization}
\label{411}
The true solution $X_t$ satisfies
\begin{equation}
    X_{t_{n+1}} = X_{t_n} + \int_{t_n}^{t_{n+1}} b(s,X_s,\mathcal{L}(X_s)) ds + \int_{t_n}^{t_{n+1}} \sigma(s,X_s,\mathcal{L}(X_s)) \mathrm{d}W_s.\nonumber
\end{equation}

Numerical solution $X_{t_{n+1}}^\textit{num}$ satisfies
\begin{equation}
    {t_{n+1}}^\textit{num} = X_{t_{n}}^\textit{num} + b\left(t_n,X_{t_{n}}^\textit{num},\mu_{t_n}\right)\Delta t + \sigma\left(t_n,X_{t_{n}}^\textit{num},\mu_{t_n}\right)\Delta W_n.\nonumber
\end{equation}

Define the error $e_n = X_{t_n}^\textit{num} - X_{t_n}^\textit{true}$, then the error satisfies the recurrence relation
\begin{equation}
\begin{aligned}
    e_{n+1} = & e_n + \left( b\left(t_n,X_{t_{n+1}}^\textit{num},\mu_{t_n}\right) - b\left(t_n,X_{tn}^{true},\mu_{t_n}\right) \right) \Delta t  \\ &+ \left( \sigma\left(t_n,X_{t_{n+1}}^\textit{num},\mu_{t_n}\right) - \sigma\left(t_n,X_{t_n}^{true},\mu_{t_n}\right) \right) \Delta W_n + R_n,
\end{aligned}
    \nonumber
\end{equation}
where $R_n$ is the higher order residual term.

Using the Lipschitz continuity of $b$ and $\sigma$, it can be shown that
\begin{equation}
    \mathbb{E}[|e_{n+1}|^p] \leq (1+C\Delta t)\mathbb{E}[|e_n|^p] + C(\Delta t)^{p/2},\nonumber
\end{equation}
and
\begin{equation}
    \mathbb{E}[|e_n|^p]=\mathcal{O}((\Delta t)^{p/2}).\nonumber
\end{equation}

Therefore, the strong error can be defined as
\begin{equation}\mathbb{E}\left[\sup_{0\leq n\leq N}|X_{t_n}^\mathrm{num}-X_{t_n}^\mathrm{true}|^p\right]^{1/p}\nonumber\end{equation}
to measure the path deviation between the numerical solution $X_{t_n}^\mathrm{num}$ and the true solution $X_{t_n}^\mathrm{true}$, where $p\geq 1$ is the order of the paradigm.

For the Euler-Maruyama method, the order of convergence of the strong error is known to be, under appropriate regularity conditions
\begin{equation}\mathbb{E}\left[\sup_{0\leq n\leq N}|X_{t_n}^{\mathrm{num}}-X_{t_n}^{\mathrm{true}}|^p\right]^{1/p}=\mathcal{O}(\Delta t^{1/2}),\nonumber\end{equation}
which is the rate of convergence of the strong error is $\mathcal{O}(\Delta t^{1/2}).$

The weak error measures the deviation of the numerical solution from the true solution in terms of the expected value and is defined as $$\left|\mathbb{E}[f(X_T^\mathrm{num})]-\mathbb{E}[f(X_T^\mathrm{true})]\right|,$$
where $f$ is a smooth test function.

For the Euler-Maruyama method, the order of convergence of the weak error under appropriate regularity conditions is
$$\left|\mathbb{E}[f(X_T^\mathrm{num})]-\mathbb{E}[f(X_T^\mathrm{true})]\right|=\mathcal{O}(\Delta t),$$
That is, the rate of convergence of the weak error is $\mathcal{O}({\Delta t}).$

\subsubsection{Approximation Errors of Probability Distributions}
\label{412}
The drift term $b(t,x,\mu_t)$ and the diffusion term $\sigma(t,x,\mu_t)$ in the MKV FBSDEs depend explicitly on the distribution $\mu_t$. In practical calculations, the distribution $\mu_t$ is usually approximated by finite samples.

Suppose $\mu_t^\mathrm{particle}$ is an empirical distribution based on $M$ independent identically distributed samples
\begin{equation}
\mu_t^\text{particle}=\frac1M\sum_{i=1}^M\delta_{X_t^{(i)}},\nonumber
\end{equation}
where $X_t^{(i)}$ is the state of the $i-$th particle.

The Wasserstein distance between the empirical distribution $\mu_t^\mathrm{particle}$ and the true distribution $\mu_t^\mathrm{true}$ satisfies the central limit theorem
\begin{equation}
    \mathbb{E}[\|\mu_t^\text{particle}-\mu_t^\text{true}\|]=\mathcal{O}(M^{-1/2}).\nonumber
\end{equation}
Assuming $ M $ particles are used to approximate the distribution, the approximation error of the distribution is
\begin{equation}
    \|\mu_t^\text{particle}-\mu_t^\text{true}\|=\mathcal{O}(M^{-1/2}),\nonumber
\end{equation}
where $M$ is the number of particles.

\subsection{Analysis of Computational Complexity}
The computational complexity of numerical methods, such as the finite element method, when directly solving the HJB-FPK system, is related to the number of mesh cells used for discretization. Let $N_h$ denote the number of mesh cells, where $h$ is the mesh size (the characteristic length of the cells), and the mesh is uniformly distributed. In a $d$-dimensional space, the number of mesh cells scales as $N_h \sim h^{-d}$, meaning that solving the HJB-FPK equations numerically requires discretization across all dimensions at each time step, with computation for each mesh cell. Assuming a time discretization step size $\Delta t$, the computational complexity is $\mathcal{O}(N_h/\Delta t)$. As the dimensionality $d$ increases, the number of mesh cells grows exponentially, making high-dimensional problems difficult to solve.

Machine learning-based methods, such as the RL-PIDL \cite{chen2023hybrid}, solve the HJB and FPK equations using neural networks. The computational complexity depends on the number of collocation points $N_p$ and training iterations per time step. Assuming fixed network architectures, the complexity is $\mathcal{O}(N_p/\Delta t)$, where $N_p$ typically scales polynomially with $d$, in contrast to the exponential scaling $N_h \sim h^{-d}$ in mesh-based methods.

In sampling-based methods, such as APAC-NET \cite{lin2021alternating} and MFGNet \cite{ruthotto2020machine}, the loss for network training is computed through interactions between representative agents. If the number of sampled agents is $N$, interactions between all pairs of agents must be calculated at each time step. Assuming a time discretization step $\Delta t$, the computational complexity of these methods is $\mathcal{O}(N^2/\Delta t)$.

In the NF-MKV Net approach, the equilibria of the MFG is estimated by sampling representative agents, tracking their evolution, and computing marginal distributions. The computational complexity consists of two parts: first, calculating the dynamics of the representative agents' states and value functions via MKV FBSDEs, and second, fitting the density distributions of the sampled agents using NF. During this process, the representative agents interact with the population density distribution. The computational complexity of estimating the dynamics is $\mathcal{O}(N/\Delta t)$, and the complexity of fitting the NF is also $\mathcal{O}(N/\Delta t)$, resulting in an overall complexity of $\mathcal{O}(N/\Delta t)$ for the NF-MKV Net approach.

The computational complexity of the NF-MKV Net approach is $\mathcal{O}(N/\Delta t)$, while that of sampling-based methods is $\mathcal{O}(N^2/\Delta t)$, and that of numerical methods for solving the full-space HJB-FPK equations is $\mathcal{O}(N_h/\Delta t)$, with machine learning-based methods scaling as $\mathcal{O}(N_p/\Delta t)$. Therefore, the NF-MKV Net approach proposed in this paper offers a computational advantage in terms of complexity.

\begin{table}[ht]
\centering
\caption{Analysis of Computational Complexity}
\label{ACC}
\centering
\begin{tabular}{cccc}
\toprule
~ & Method& Complexity & Sensitivity\\
\midrule
\multirow{2}{*}{\makecell[c]{Solving \\ HJB-FPK}}&Mess-based\cite{doi:10.1137/120882421}&$\mathcal{O}(N_h/\Delta t)$&CoD.\footnotemark($d \uparrow$)\\
~&ML-based \cite{chen2023hybrid}&$\mathcal{O}(N_p/\Delta t)$&Polynomial ($d \uparrow$)\\
\midrule
\multirow{2}{*}{\makecell[c]{Sampling\\Points}} & APAC-NET \cite{lin2021alternating}& $\mathcal{O}(N^2/\Delta t)$ &Polynomial ($N \uparrow$)\\
~ & NF-MKV Net& $\mathcal{O}(N/\Delta t)$ & Linear ($N \uparrow$)\\
\bottomrule

\end{tabular}
\end{table}
\footnotetext{CoD.: Curse of Dimensionality.}

\section{Numerical Experiment}
\label{S5}
We apply NF-MKV Net to MFG instances and present the numerical results in two parts. The first part \ref{51} demonstrates NF-MKV Net as an effective approach for solving MFG equilibria involving density distributions. The second part \ref{52} highlights the accuracy of NF-MKV Net in comparison to other algorithms.

\subsection{Solving MFG with NF-MKV Net}
\label{51}
This section presents three examples of solving MFG using NF-MKV Net, demonstrating its applicability to traffic flow problems, low- and high-dimensional crowd motion problems, scenarios with obstacles and with terminal value function.

\subsubsection{Example 1: MFG Traffic Flow Control}
A series of numerical experiments in MFG Traffic Flow Control explore the dynamics of MFG, focusing on autonomous vehicles navigating a circular road network. The traffic flow scenario is formulated as an MFG problem involving density distribution and the value function.

The initial density is defined on the ring road, where the state $x$ represents the AVs' position. The state transfer function is $\mathrm{d}x=b\mathrm{d}t+\sigma \mathrm{d}W_t$, and the process constraint is $f(x)=\frac{1}{2}(1-\mu(x) -b)^2$. The Hamiltonian is defined as $H(x,p,t)=f(p,\mu)+pu_x$, leading to the optimal control $u^*=\arg \min_p(f(p,\mu)+pu_x)$. In the finite time domain problem, the terminal value function $u_T$ of the AVs system is constrained at $t=T$. It is assumed that AVs have no preference for their locations at time $T$, i.e., $u(x, T) = 0$. In the MFG traffic flow problem, the terminal value function $u_T$ can be solved explicitly as $\mu_T(x)=1,\forall x\in (0,1)$, satisfying the model's assumptions.

Without loss of generality, we define the time interval as $[0,1]$ and set the initial density $\mu_0$ at $t=0$. To verify the volumetric invariance of the density distribution discussed in our study, we selected initial density functions satisfying $\int_0^1 \mu_0(x)\mathrm{d}x$. Four different initial densities were selected, each with a distinct diffusion coefficient $\sigma$ for the Wiener process. NF-MKV Net was then employed to solve for equilibria, verifying the proposed algorithm's applicability.

\begin{figure}[h]
\begin{center}
\includegraphics[width=\linewidth]{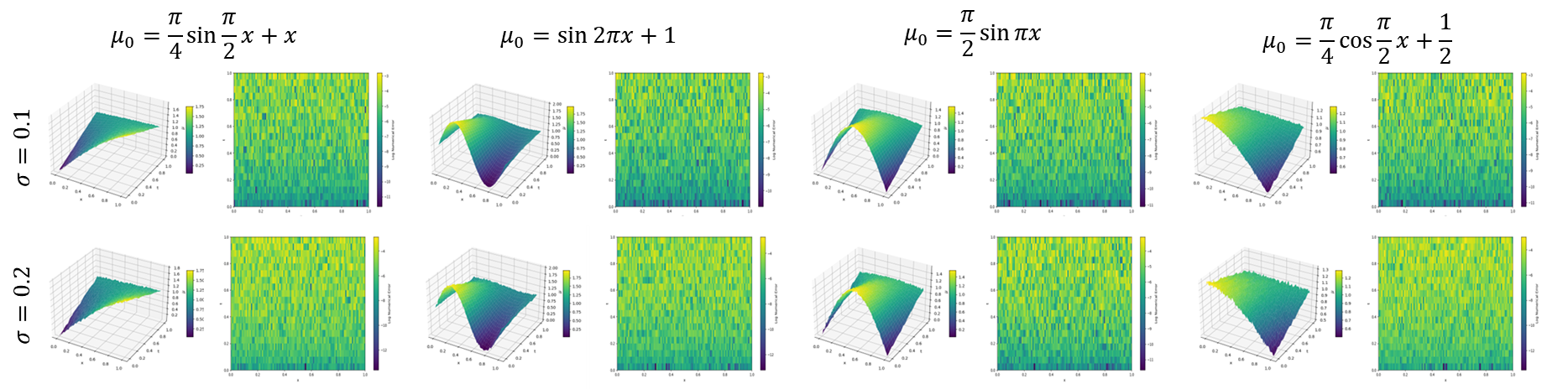}
\end{center}
\caption{NF-MKV Net solutions and numerical error of MFG traffic flow with various initial density distribution and diffusion coefficients}
\label{example-1d}
\end{figure}

Results in Figure \ref{example-1d} indicate that the agent distribution, regardless of initial density $\mu_0$ or drift term $\sigma$, converges to the equilibria $\mu(x, T)=1$. Comparing NF-MKV Net with the numerical solution shows errors below $10^{-3}$, demonstrating the algorithm's effectiveness in solving traffic flow problems. The accuracy of the numerical method for solving the MFG equilibria of one-dimensional traffic flow can reach within $10^{-5}$ \cite{achdou2012mean}, so the effectiveness of the proposed MFG equilibria solving method can be verified by comparing it with the numerical method. The results illustrate the evolution of the density distribution $\mu(x)$ over time $t$ while the square diagram shows the the $\log$ errors compared to the noise-free numerical method that
\begin{equation*}
    \varepsilon = \log {\frac{\|\mu_t^{\text{Net}}(x)-\hat{\mu}_t(x)\|}{\hat{\mu}_t(x)}},
\end{equation*}
where $\mu_t^{\text{Net}}(x)$ and $\hat{\mu}_t(x)$ denote the density distributions of state $x$ at time $t$, calculated by the NF-MKV Net and the numerical method, respectively. 

Simulation results demonstrate that NF-MKV Net is capable of addressing traffic flow problems, while numerical results indicate the approach's high accuracy. These results underscore the applicability and precision of NF-MKV Net in solving density-coupled MFG equilibria.

\subsubsection{Example 2: MFG Crowd Motion}
In this example, a dynamically formulated MFG problem, the Crowd Motion problem, is constructed in dimensions $d=2$ and $d=50$ to demonstrate the applicability of NF-MKV Net. We set the problems as in equations \ref{mfg} with parameter
\begin{equation}
\begin{aligned}
    &f(x,\mu) = \int _{\mathbb{R}^d} e^{-|x - \hat{x}|^2} \mu(\hat{x})d\hat{x}, \quad &&H(x,p,t) = |p|^2 + f(x,\mu,t), \\
    &\mu(x,0)= \mu_0(x), &&u(x,T)=e^{|x - x_T|^2}.
\end{aligned}
\label{cm2d}
\end{equation}

$\mathbf{d=2}$ \textbf{Crowd Motion.} Here, $\sigma = \sqrt{2}$ is used with 100 time steps in the dynamics process, and the initial distribution is set as $\mu_0(x)=\mathcal{N}((-2,0), \allowbreak (0.5^2, 0.5^2))$. To reach the goal point, we set $x_T=(2,0)$ which means the terminal value function is $g(x,\mu(\cdot))=e^{|x - (2,0)|^2}$. With these settings, NF-MKV Net trains the MFG model associated with the dynamic system.

\begin{figure}[h]
\begin{center}
\includegraphics[width=\linewidth]{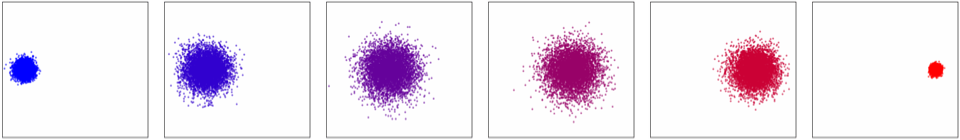}
\end{center}
\caption{2-dimensional crowd motion dynamics flow}
\label{2-d}
\end{figure}

$\mathbf{d=50}$ \textbf{Crowd Motion.} High-dimensional methods adopt the same settings, similar to the 2-dimensional case, as in equations \ref{mfg} and equations \ref{cm2d}. In contrast, high-dimensional methods handle agent states and controls in $\mathbb{R}^{50}$, along with density distributions in $\mathcal{L}(\mathbb{R}^{50})$. So Our initial density $\mu_0(x)$ is a Gaussian centered at $(-2,-2,0,\cdots ,0)$ and terminal value function $g(x,\mu(\cdot))=e^{|x - (2,2,0,\cdots , 0)|^2}$. With these settings, NF-MKV Net trains the MFG model associated with the dynamic system. Results are visualized in the first two dimensions by summing projections from higher dimensions onto these two dimensions.

\begin{figure}[h]
\begin{center}
\includegraphics[width=\linewidth]{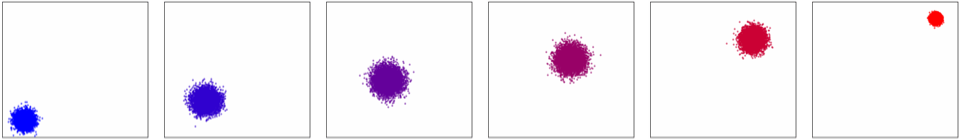}
\end{center}
\caption{50-dimensional crowd motion dynamics flow}
\label{50-d}
\end{figure}

The trajectories of $500$ points are shown in Figure \ref{2-d} for the 2-dimensional case and Figure \ref{50-d} for the 50-dimensional case. NF-MKV Net effectively transforms the initial Gaussian density into the terminal value function along a nearly straight trajectory, while ensuring crowd deformation and inter-group collision avoidance. This behavior remains consistent as the dimensionality increases.

Simulation results show that NF-MKV Net is capable of handling both low- and high-dimensional crowd motion problems, demonstrating its applicability in solving MFG equilibria for both low- and high-dimensional cases.

\subsubsection{Example 3: MFG Crowd Motion with obstacle}
This experiment considers an MFG problem with complex process interaction costs. Following the general setting in equations \ref{mfg} and equations \ref{cm2d}. We set  $\sigma = \sqrt{2}$ with 20 time steps in the dynamics process and set initial distribution as $\mu_0(x)=\mathcal{N}((-4,0),(0.1^2,0.1^2))$. To reach the goal point, we set $x_T=(4,0)$ which means the terminal value function is $g(x,\mu(\cdot)) = e^{|x - (2,2)|^2}$. During the dynamics, the system optimizes the process loss $f$ defined in equation \ref{ff}. The problem involves transforming the initial Gaussian density to a new location while minimizing terminal value function, avoiding congestion, and bypassing an obstacle at $x_o = (0, 0)$ with a safety radius $s_{\text{safe}} = 1$. With these settings, NF-MKV Net successfully trains the MFG model associated with the dynamics system.

\begin{figure}[h]
\begin{center}
\includegraphics[width=\linewidth]{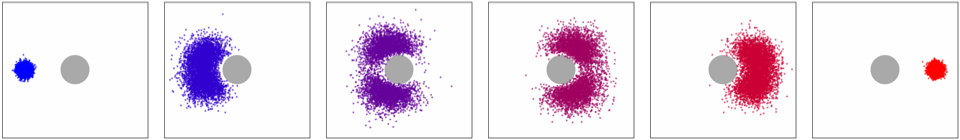}
\end{center}
\caption{2-dimensional crowd motion dynamics flow with an obstacle.}
\label{2dob}
\end{figure}

The results, shown in figure \ref{2dob}, demonstrate that NF-MKV Net successfully transforms the initial Gaussian density into the desired terminal value function along an optimized trajectory, ensuring crowd deformation, inter-group collision avoidance, and obstacle avoidance. Simulation results demonstrate the applicability of NF-MKV Net in solving non-linear MFG equilibria.

\subsubsection{Example 4: MFG Crowd Motion without terminal distribution}

To verify that our approach differs from the optimal transport approach with MFG constraints, we conducted an experiment on MFG-based crowd motion, where the terminal state density is not pre-specified. This experiment involves solving the MFG equations under initial distribution $\mu_0(x)=\mathcal{N}((-4,0),(0.1^2,0.1^2))$ and the loss of terminal value function constraints $g((x_1,x_2),\mu_T(\cdot))=e^{|x_1 - 4|^2}$, which, in practical scenarios, can correspond to enforcing terminal value function constraints that require agents to leave a specific region. This setting is naturally formulated as an MFG problem rather than an OT problem with terminal density matching. This distinction further clarifies the differences between our approach and the optimal transport approach with MFG constraints.

\begin{figure}[h]
\begin{center}
\includegraphics[width=\linewidth]{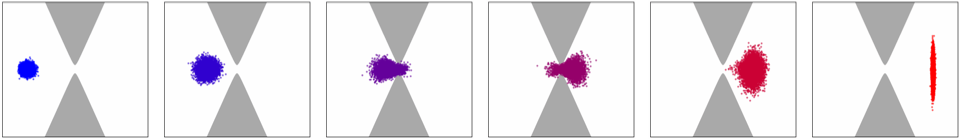}
\end{center}
\caption{2-dimensional crowd motion dynamics flow with terminal value function rather than distribution.}
\label{line}
\end{figure}

The results in figure \ref{line} show that NF-MKV Net transforms the initial Gaussian density into the desired terminal value function along an optimized trajectory, ensuring crowd deformation, inter-group collision avoidance, and obstacle avoidance.

Notably, in this experiment, the distribution along the $x_2$ component ($y-$axis) was not specified. The results show that agents disperse along the $y-$axis in a manner determined by the diffusion coefficient. Consequently, the density distribution cannot be predetermined when setting up the experiment. This illustrates the difference between our approach and OT with MFG constraints.

\subsection{Comparison with other Methods}
\label{52}
To verify volumetric invariance and time continuity, we compare NF-MKV Net with existing MFG solving methods, including the distribution-based \textbf{RL-PIDL} method \cite{chen2023hybrid} and the high-dimensional neural network-based \textbf{APAC-Net} \cite{lin2021alternating}.

\textbf{Distribution volumetric-invariance.} We implement an approximated integral over the dynamics region, widely used in density estimation \cite{WOS:000378444200012}. By generating a mesh over a specified area, the numerical integration of a specified probability distribution over that area is computed, and the return value should be close to 1. This method verifies the validity of the distribution. Since the approximated integral is a mesh-based method, it can only be used in low-dimensional problems. So, when we approximate integral in $d=50$ crowd motion, we use the same process as when showing the high-dimensional trajectories from the experiment above, that is, a projection-like method that accumulates the density distribution function on the other components over the first two components and estimates the density in a 2-dimensional region.

\textbf{Agents states time-continuity.} The Wasserstein distance is generally chosen as the metric for the difference between two density distributions. The method has been used to assess the difference between distributions \cite{lauriere2022scalable}. Even without an explicit probability density function, the Wasserstein distance ($W$-dis) can be computed by optimization methods as long as it can be sampled from both distributions.

\begin{figure}[ht]

  \centering
  \begin{minipage}[b]{0.45\textwidth}
    \includegraphics[width=\textwidth]{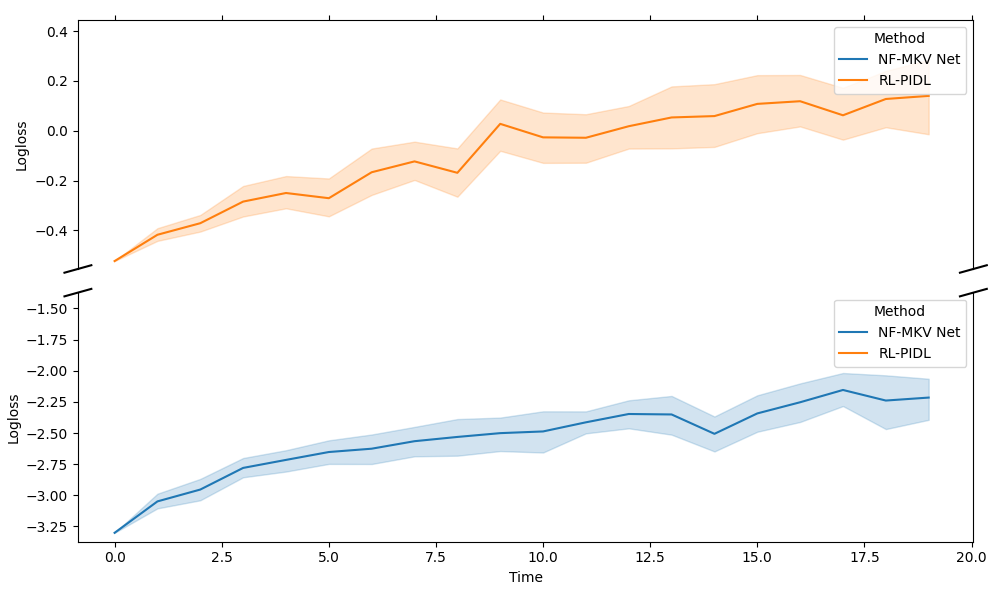}
  \end{minipage}
  \hfill
  \begin{minipage}[b]{0.45\textwidth}
    \includegraphics[width=\textwidth]{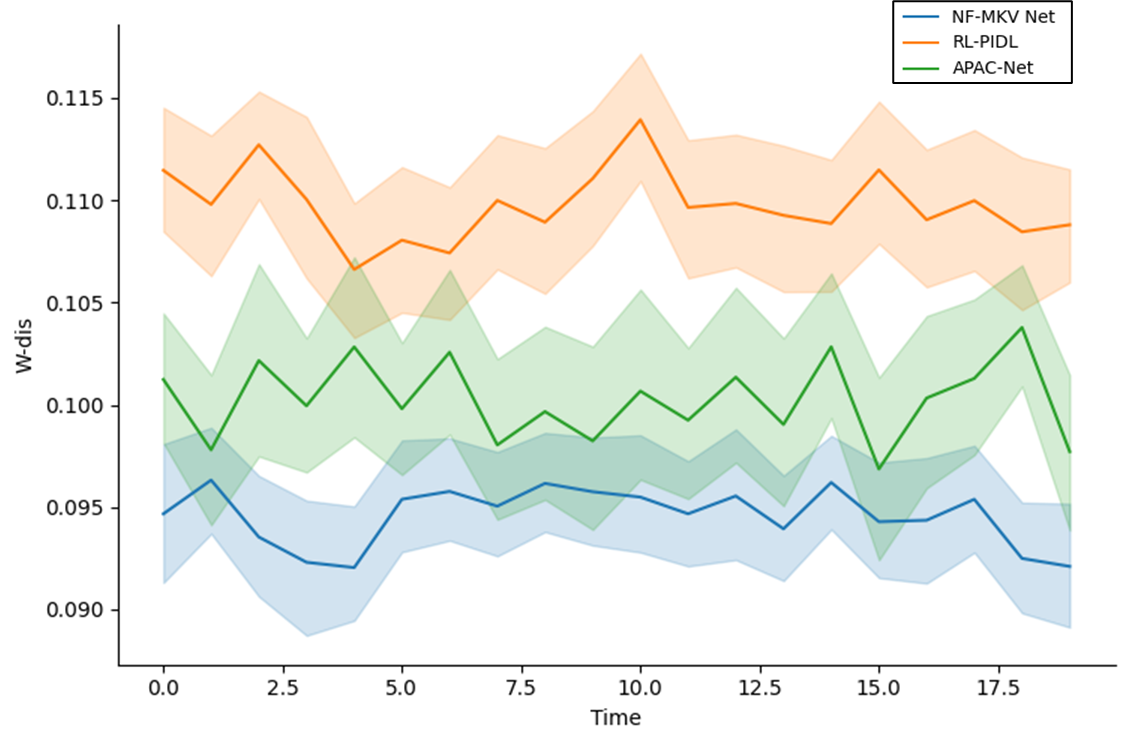}
  \end{minipage}
  
  \caption{Comparison results in time steps. Left: $\log$ of integration of distribution difference from $1$ in traffic flow ($d=1$). Right: Wasserstein distance of distribution between time steps in crowd motion ($d=2$).}
  \label{123}
\end{figure}

We set the comparison experiments in traffic flow ($d=1$) and crowd motion ($d=2$ and $50$). The comparison results are shown in table \ref{comp} and figure \ref{123}.

\begin{table}[ht]
\caption{Numerical Comparison with Other Methods}
\label{comp}
\centering
\begin{tabular}{ccccc}
\toprule
~ &Aspects  &Ours
&RL-PIDL\cite{chen2023hybrid} &APAC-NET\cite{lin2021alternating}
\\ \midrule
\multirow{2}{*}{\makecell[c]{Traffic \\ Flow\\$(d=1)$}}&\makecell[l]{$\log $ of $\mu$ integral\\ difference from $1$}     & $-2.67$ & $-0.21$ & \makecell[c]{/\ \\(sample-based)}\\  

 ~ & \makecell[l]{$W$-dis of $\mu_t$ bet-\\ween time steps} & $0.044$ &$0.052$ & $0.047$\\

\midrule

\multirow{2}{*}{\makecell[c]{Crowd \\ Motion\\$(d=2)$}}&\makecell[l]{$\log $ of $\mu$ integral\\ difference from $1$}     & $-2.32$ & $-0.13$ & \makecell[c]{ /\ \\(sample-based)}\\  

 ~ & \makecell[l]{$W$-dis of $\mu_t$ bet-\\ween time steps} & $0.096$ & $0.108$ & $0.101$\\

\midrule

\multirow{2}{*}{\makecell[c]{Crowd\\Motion\\$(d=50)$}}&\makecell[l]{$\log $ of $\mu$ integral\\ difference from $1$}     & $-1.06$ & $0.26$ & \makecell[c]{/\ \\(sample-based)}\\ 

 ~ & \makecell[l]{$W$-dis of $\mu_t$ bet-\\ween time steps} & $2.27$ & $3.49$ & $2.85$\\

\bottomrule
\end{tabular}

\end{table}

In the results of \textbf{density integral errors}, the NF-MKV Net approach exhibits an error of less than $10^{-2}$ between the density integral over the feasible domain and 1, whereas the errors of the other methods exceed $10^{-1}$. In high-dimensional scenarios, the error for NF-MKV Net is $10^{-1.06}$, while the errors for the other methods surpass $1$. This analysis indicates that,  in high-dimensional scenarios, the relative error for the other methods exceeds $100\%$, while the relative error for NF-MKV Net is approximately $10\%$. These findings demonstrate that the NF-MKV Net approach maintains the property of volumetric invariance of the distribution integral of the group density throughout the time evolution during the solution process.

In the high-dimensional results of the \textbf{average Wasserstein distance} between adjacent time steps, the average value of individual trajectory density solved by the NF-MKV Net approach is $2.27$, while the values for the other two methods are $3.49$ and $2.85$, respectively. This highlights that among the three methods, NF-MKV Net achieves the smallest average distance between adjacent time steps, indicating smooth evolution and good agents states time-continuity.

Numerical results indicate that NF-MKV Net excels in density volumetric invariance and agent state time-continuity, making it suitable for solving MFG problems involving density distributions.

\section{Conclusion}
\label{S6}
This paper presents the NF-MKV Net, a neural network-based approach for solving high-dimensional MFG equilibria, formulated as fixed-point problems of MKV FBSDEs. The approach integrates process-regularized NF with state-policy-connected time-series Neural Networks, effectively modeling the evolution of density distributions and value functions while ensuring adherence to of high-dimensional, density-coupling and time-continuity. The approach was theoretically analyzed, focusing on time discretization, probability distribution approximations errors and computational complexity, which demonstrated that the NF-MKV Net can achieve high accuracy with the theoretical analysis in both low- and high-dimensional problems. Through numerical experiments, including traffic flow, crowd motion, and obstacle avoidance tasks, NF-MKV Net was shown to not only solve high-dimensional, density-coupling MFG equilibria but also maintain the consistency of density evolution, as evidenced by the low error rates in both distribution volumetric invariance and agent state time-continuity. The results obtained across various scenarios underline the potential of this approach in real-world applications, offering a novel tool for solving high-dimensional MFG equilibria with constraint.

\section{CRediT authorship contribution statement}
\textbf{Jinwei Liu}: Conceptualization, Formal analysis, Methodology, Software, Visualization, Writing – original draft. \textbf{Lu Ren}: Conceptualization, Formal analysis, Writing – original draft. \textbf{Wang Yao}: Supervision
Writing – review \& editing. \textbf{Xiao Zhang}: Funding acquisition, Supervision, Validation.
\section{Declaration of competing interest}
Authors declare no competing financial interest.

\section{Acknowledgments}
This work was supported by the National Science and Technology Major Project (Grant No. 2022ZD0116401) and the Research Funding of Hangzhou International Innovation Institute of Beihang University (Grant No. 2024KQ161).

\section{Data availability}
No data was used for the research described in the article.

\newpage

\bibliographystyle{elsarticle-num}
\bibliography{refe.bib}

\end{document}